\documentclass[11pt,a4paper]{article}
\usepackage{authblk}
\usepackage[hyperref]{acl2021}
\usepackage{times}
\usepackage{latexsym}
\usepackage{color}
\usepackage{hyperref}
\usepackage{booktabs}
\usepackage{graphicx}\usepackage{amsmath}
\usepackage{stfloats}
\usepackage{lipsum}
\usepackage{tikz}
\usepackage{cleveref}

\crefformat{section}{\S#2#1#3} % see manual of cleveref, section 8.2.1
\crefformat{subsection}{\S#2#1#3}
\crefformat{subsubsection}{\S#2#1#3}

\DeclareGraphicsExtensions{.PNG}

\newcommand\encircle[1]{%
  \tikz[baseline=(X.base)] 
    \node (X) [draw, shape=circle, inner sep=0] {\strut #1};}

\usepackage{microtype}

\aclfinalcopy % Uncomment this line for the final submission
\title{IntelliCAT: Intelligent Machine Translation Post-Editing with Quality Estimation and Translation Suggestion}

\renewcommand*{\Affilfont}{\normalsize\normalfont}
\newsavebox\affbox
\author[1]{Dongjun Lee}
\author[1]{Junhyeong Ahn}
\author[1]{Heesoo Park}
\author[2]{Jaemin Jo}
\affil[1]{%
{\Affilfont Bering Lab, Republic of Korea}}
\affil[2]{%
{\Affilfont Sungkyunkwan University, Republic of Korea}}
\affil[ ]{\textit {\{djlee, rkdrnf, heesoo.park\}@beringlab.com, jmjo@skku.edu}}

\date{}

\begin{document}
\maketitle
\begin{abstract}
We present IntelliCAT, an interactive translation interface with neural models that streamline the post-editing process on machine translation output. 
We leverage two quality estimation (QE) models at different granularities: sentence-level QE, to predict the quality of each machine-translated sentence, and word-level QE, to locate the parts of the machine-translated sentence that need correction.
Additionally, we introduce a novel translation suggestion model conditioned on both the left and right contexts, providing alternatives for specific words or phrases for correction.
Finally, with word alignments, IntelliCAT automatically preserves the original document's styles in the translated document.
The experimental results show that post-editing based on the proposed QE and translation suggestions can significantly improve translation quality. 
Furthermore, a user study reveals that three features provided in IntelliCAT significantly accelerate the post-editing task, achieving a 52.9\% speedup in translation time compared to translating from scratch.
The interface is publicly available at \href{https://intellicat.beringlab.com/}{https://intellicat.beringlab.com/}.
\end{abstract}

\section{Introduction}

Existing computer-aided translation (CAT) tools incorporate machine translation (MT) in two ways: post-editing (PE) or interactive translation prediction (ITP).
PE tools \citep{matecat, catalog_online} provide a machine-translated document and ask the translator to edit incorrect parts.
By contrast, ITP tools \citep{casmacat, lilt, inmt} aim to provide translation suggestions for the next word or phrase given a partial input from the translator.
A recent study with human translators revealed that PE was 18.7\% faster than ITP in terms of translation time \citep{green2014human} and required fewer edits \citep{do2020comparing}. However, many translators still prefer ITP over PE because of (1) high cognitive loads \citep{koehn2009process} and (2) the lack of subsegment MT suggestions \citep{moorkens2017assessing} in PE.

In this paper, we introduce IntelliCAT\footnote{A demonstration video is available at \href{https://youtu.be/mDmbdrQE9tc}{https://youtu.be/mDmbdrQE9tc}}, a hybrid CAT interface designed to provide PE-level efficiency while retaining the advantages of ITP, such as subsegment translation suggestions. To mitigate the cognitive loads of human translators, IntelliCAT aims to automate common post-editing tasks by introducing three intelligent features: (1) quality estimation, (2) translation suggestion, and (3) word alignment.

Quality estimation (QE) is the task of estimating the quality of MT output without reference translations \citep{qe-20}. We integrate QE into the CAT interface so that the human translator can easily identify which machine-translated sentences and which parts of the sentences require corrections.
Furthermore, for words that require post-editing, our interface suggests possible translations to reduce the translators' cognitive load.
Finally, based on word alignments, the interface aligns the source and translated documents in terms of formatting by transferring the styles applied in the source document (e.g., bold, hyperlink, footnote, equation) to the translated document to minimize the post-editing time.
Our contributions are:
\begin{itemize}
    \item We integrate state-of-the-art sentence-level and word-level QE \citep{my-qe} techniques into an interactive CAT tool, IntelliCAT.
    \item We introduce a novel words and phrases suggestion model, which is conditioned on both the left and right contexts, based on XLM-RoBERTa \citep{xlm-r}. The model is fine-tuned with a modified translation language modeling (TLM) objective \citep{xlm}.
    \item We conduct quantitative experiments and a user study to evaluate IntelliCAT.
\end{itemize}

The experimental results on the WMT 2020 English-German QE dataset show that post-editing with the proposed QE and translation suggestion models could significantly improve the translation quality ($-$6.01 TER and $+$6.15 BLEU).
Moreover, the user study shows that the three features provided by IntelliCAT significantly reduce post-editing time (19.2\%), which led to a 52.6\% reduction in translation time compared to translating from scratch.
Finally, translators evaluate our interface to be highly effective, with a SUS score of 88.61.

\begin{figure*}
	\centering\includegraphics[scale=0.36]{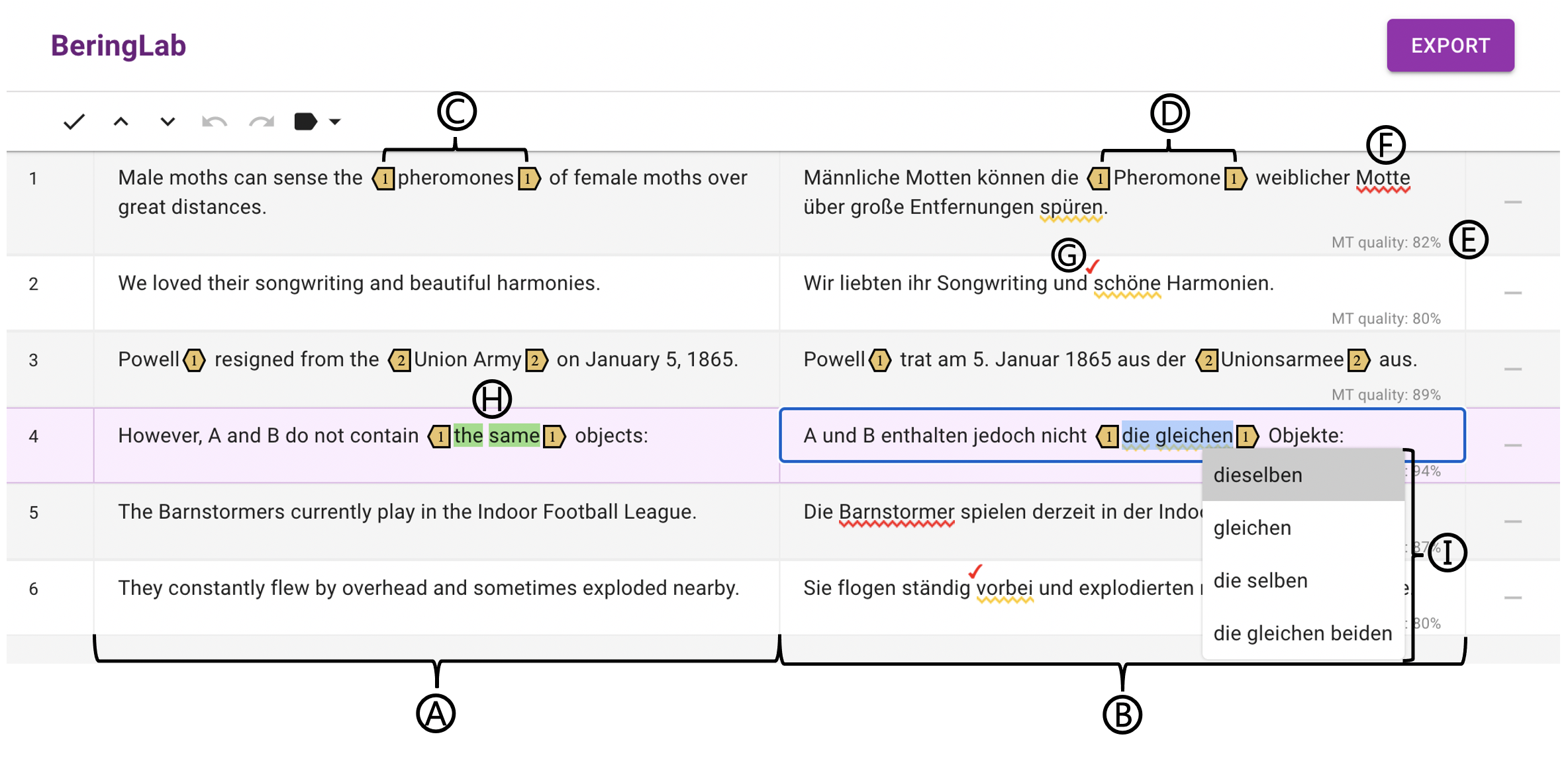}
	\caption{\label{screenshot-overview} The IntelliCAT Interface.
	 After a document (i.e., an MS Word file) is uploaded, \encircle{A} sentences from the original document (source) and \encircle{B} the initial MT output for each sentence (target) are shown side-by-side. \encircle{C} Formatting tags indicate where a specific style (identified by an integer style id) is applied and \encircle{D} are automatically inserted at the proper position of the MT output based on word alignments. \encircle{E} The interface shows the quality of each machine-translated sentence based on sentence-level QE. \encircle{F} Potentially incorrect words and \encircle{G} locations of missing words are highlighted based on word-level QE.
	 When the user selects a sequence of words in the MT output, \encircle{H} the corresponding words in the source sentence are highlighted with a heat map, and \encircle{I} up to five alternative translations are recommended.
}
\end{figure*}

\section{Related Work}

\paragraph{CAT Tool and Post-Editing}

In the localization industry, the use of CAT tools is a common practice for professional translators \citep{van2015recommendations}.
As MT has improved substantially in recent years, approaches incorporating MT into CAT tools have been actively researched \citep{casmacat, matecat, inmt, mmpe}.
One of the approaches is post-editing in which the translator is provided with a machine-translated draft and asked to improve the draft.
Recent studies demonstrate that post-editing MT output not only improves translation productivity but also reduces translation errors \citep{green2013efficacy, aranberri2014comparison, toral2018post}.

\paragraph{Translation Suggestion}
Translation suggestions from interactive translation prediction (ITP) \citep{casmacat, inmt, intellingo} are conditioned only on the left context of the word to be inserted. Therefore, ITP has intrinsic limitations in post-editing tasks where the complete sentence is presented, and the right context of the words that need correction should also be considered.
We propose a novel translation suggestion model in which suggestions are conditioned on both the left and right contexts of the words or phrases to be modified or inserted to provide more accurate suggestions when post-editing the complete sentence.

\paragraph{Cross-Lingual Language Model}
Cross-lingual language models (XLMs), which are language models pre-trained in multiple languages, have led to advances in MT \citep{xlm} and related tasks such as QE \citep{my-qe}, automatic post-editing \citep{ape-1, ape-2}, and parallel corpus filtering \citep{lo2020improving}. Accordingly, our QE and translation suggestion models are trained on top of XLM-R \citep{xlm-r}, an XLM that shows state-of-the-art performance for a wide range of cross-lingual tasks.
To the best of our knowledge, IntelliCAT is the first CAT interface that leverages XLM to assist human post-editing for MT outputs.

\section{System Description}
\subsection{Overview}

IntelliCAT is a web-based interactive interface for post-editing MT outputs (\autoref{screenshot-overview}).
Once loaded, it shows two documents side-by-side: the uploaded original document (an MS Word file) on the left and the machine-translated document on the right. 
Each document is displayed as a list of sentences with \textit{formatting tags} inserted, tags that show the style of the original document, including text styles (e.g., bold, italic, or hyperlinked) and inline contents (e.g., a media element or an equation).

The user can post-edit MT outputs on the right using the following three features: (1) sentence-level and word-level QE, (2) word or phrase suggestion, and (3) automatic tagging based on word alignments.
The sentence-level QE shows the estimated MT quality for each sentence, and word-level QE highlights the parts of each machine-translated sentence that need correction. When the user selects a specific word or phrase, the top-$5$ recommended alternatives appear below, allowing the user to replace the selected words or insert a new word. Finally, the system automatically captures the original document style and inserts formatting tags in machine-translated sentences at the appropriate locations.
After post-editing, the user can click on the export button to download the translated document with the original style preserved.
A sample document and its translated document without human post-editing is presented in Appendix~\ref{sec:appendix-a}.

\subsection{Machine Translation}
\label{sec:nmt}
Our system provides MT for each sentence in the input document. We build our NMT model based on Transformer \citep{transformer} using OpenNMT-py \citep{opennmt}.
As training data, the English-German parallel corpus provided in the  2020 News Translation Task \citep{wmt20-findings} is used.
We use unigram-LM-based subword segmentation \citep{subword-regularization} with a vocabulary size of 32K for English and German, respectively, and the remaining hyperparameters follow the base model of \citet{transformer}.

\subsection{Quality Estimation}
Quality estimation (QE) is the task of estimating the quality of the MT output, given only the source text \citep{qe-2019}. We estimate the quality at two different granularities: sentence and word levels.
Sentence-level QE aims to predict the human translation error rate (HTER) \citep{ter-tool} of a machine-translated sentence, which measures the required amount of human editing to fix the the machine-translated sentence. 
By contrast, word-level QE aims to predict whether each word in the MT output is \texttt{OK} or \texttt{BAD} and whether there are missing words between each word. 

\autoref{screenshot-overview} demonstrates the use of QE in our interface. Based on the sentence-level QE, we show the MT quality for each machine-translated sentence computed as $1-(predicted \: HTER)$. In addition, based on word-level QE, we show words that need to be corrected (with red or yellow underlines) or locations for missing words (with red or yellow checkmarks).
To display the confidence of word-level QE predictions, we encode the predicted probability of the color of underlines and checkmarks (yellow for $P_{BAD}>0.5$ and red for $P_{BAD}>0.8$).

For QE training, we use a two-phase cross-lingual language model fine-tuning approach following \citet{my-qe}, which showed the state-of-the-art performance on the WMT 2020 QE Shared Task \citep{qe-20}. We fine-tune XLM-RoBERTa \citep{xlm-r} with a few additional parameters to jointly train sentence-level and word-level QEs. We train our model in two phases. First, we pre-train the model with a large artificially generated QE dataset based on a parallel corpus. Subsequently, we fine-tune the model with the WMT 2020 English-German QE dataset \citep{qe-20}, which consists of 7,000 triplets consisting of source, MT, and post-edited sentences.

\subsection{Translation Suggestion}
As shown in \autoref{screenshot-overview}, when the user selects a specific word or phrase to modify or presses a hotkey (\texttt{ALT+s}) between words to insert a missing word, the system suggests the top-$5$ alternatives based on fine-tuned XLM-R.

\paragraph{XLM-R Fine-Tuning}
For translation suggestion, we fine-tune XLM-R with a modified translation language modeling (TLM) objective \citep{xlm}, which is designed to better predict the masked spans of text in the translation.
Following \citet{xlm}, we tokenize source (English) and target (German) sentences with the shared BPE model \citep{bpe}, and concatenate the source and target tokens with a separation token (\texttt{</s>}).
Unlike the TLM objective of \citet{xlm}, which randomly masked tokens in both the source and target sentences, we only mask tokens in target sentences since the complete source sentence is always given in the translation task. We randomly replace $p$\% ($p \in [15, 20, 25]$) of the BPE tokens in the target sentences by \texttt{<mask>} tokens and train the model to predict the actual tokens for the masks. In addition, motivated by SpanBERT \citep{spanbert}, we always mask complete words instead of sub-word tokens since translation suggestion requires predictions of complete words. As training data, we use the same parallel corpus that is used for MT training.

\paragraph{Inference} 

To suggest alternative translations for the selected sequence of words, we first replace it with multiple \texttt{<mask>} tokens.
The alternative translations may consist of sub-word tokens of varying lengths. Hence, we generate $m$ inputs, where $m$ denotes the maximum number of masks, and in the $i^{th}$ input ($i \in [1, ..., m]$), the selected sequence is replaced with $i$ consecutive \texttt{<mask>} tokens. In other words, we track all cases in which alternative translations consist of $1$ to $m$ sub-word tokens.
Then, each input is fed into the fine-tuned XLM-R, and \texttt{<mask>} tokens are iteratively replaced by the predicted tokens from left to right. In each iteration, we use a beam search with a beam size $k$ to generate the top-$k$ candidates. 
Finally, all mask prediction results from $m$ inputs are sorted based on probability, and the top-$k$ results are shown to the user.

\subsection{Word Alignment and Automatic Formatting}

To obtain word alignments, we jointly train the NMT model (\cref{sec:nmt}) to produce both translations and alignments following \citet{word-alignment-jointly-learning}. One attention head on the Transformer's penultimate layer is supervised with an alignment loss to learn the alignments. We use Giza++ \citep{giza} alignments as the guided labels for the training. 
As sub-word segmentation is used to train the NMT model, we convert the sub-word-level alignments back to the word-level. We consider each target word to be aligned with a source word if any of the target sub-words is aligned with the source sub-words.

We provide two features based on word alignment information. 
First, when the user selects a specific word or phrase in the machine-translated sentence, the corresponding words or phrases in the source sentence are highlighted using a heatmap.
Second, formatting tags are automatically inserted at the appropriate locations in the machine-translated sentences. We use two types of tags to represent the formatting of the document: paired tags and unpaired tags. 
Paired tags represent styles applied across a section of text (e.g., bold or italic). To retain the style applied in the source sentence to the MT, we identify the source word with the highest alignment score for each target word and apply the the corresponding source word's style to the target word.
By contrast, unpaired tags represent inline non-text contents such as media elements and equations. To automatically insert an unpaired tag in the MT, we identify the target word with the highest alignment score with the source word right before the tag and insert the corresponding tag after the target word.

\begin{table*}[h!]
    \centering
    \small
    \begin{tabular}{lcc|cc}
    \toprule
         & \multicolumn{2}{c}{(With Predicted QE)} & \multicolumn{2}{c}{(With Oracle QE)} \\
         Model & TER$\downarrow$ & BLEU$\uparrow$ & TER$\downarrow$ & BLEU$\uparrow$ \\
    \midrule
    Baseline (MT) & 31.37 & 50.37 & 31.37 & 50.37 \\
    \midrule
    XLM-R \\ \citep{xlm-r} & & & & \\
    \enspace \quad Top-1 & 30.28 (-1.09) & 50.78 (+0.41) & 26.57 (-4.80) & 56.02(+5.65) \\
    \enspace \quad Top-3 & 29.47 (-1.90) & 50.89 (+0.52) & 24.10 (-7.27) & 60.28 (+9.91) \\
    \enspace \quad Top-5 & 28.75 (-2.62) & 51.85 (+1.48) & 22.78 (-8.59) & 62.40 (+12.03) \\
    \midrule
    \multicolumn{5}{l}{Proposed} \\
    \enspace \quad Top-1 & \textbf{29.04 (-2.33)} & \textbf{51.93 (+1.56)} & \textbf{24.26 (-7.11)} & \textbf{59.38 (+9.01)} \\
    \enspace \quad Top-3 & 26.69 (-4.68) & 54.70 (+4.33) & 19.08 (-12.29) & 67.51 (+17.14) \\
    \enspace \quad Top-5 & 25.36 (-6.01) & 56.52 (+6.15) & 17.30 (-14.07) & 70.50 (+20.13) \\
    
    \bottomrule
    \end{tabular}
    \caption{TER and BLEU for machine-translated sentences (Baseline) and post-edited sentences (XLM-R and Proposed) based on word-level QE and translation suggestion.}
    \label{table:model-eval}
\end{table*}

\section{Experiments}

\subsection{Model Evaluation}

\paragraph{Experimental Setup}

To evaluate the performance of translation suggestions, we measure MT quality improvement when a sentence is corrected with the suggested words or phrases.
We introduce two selection conditions (\textbf{Oracle QE} and \textbf{Predicted QE}) and two suggestion methods (\textbf{XLM-R} and \textbf{Proposed}).
The selection conditions locate the words that need to be corrected in a sentence; in \textbf{Oracle QE} condition, the ground truth word-level QE label is used as a baseline, and in \textbf{Predicted QE} condition, our word-level QE model is used to identify the target words.
The suggestion methods determine the words that the selected words should be replaced with.
We test two suggestion models, the pre-trained XLM-R\footnote{\href{https://pytext.readthedocs.io/en/master/xlm_r.html}{https://pytext.readthedocs.io/en/master/xlm\textunderscore r.html}} and the proposed model, fine-tuned with the modified TLM objective, with three different suggestion sizes: top-1, top-3, and top-5.

Each of the QE and translation suggestion models was trained using two Tesla V100 GPUs. As an evaluation dataset, we use the WMT 2020 English-German QE \textit{dev} dataset \citep{qe-20}. As evaluation metrics, we use the translation error rate (TER) \citep{ter-tool} and BLEU \citep{bleu}.

\paragraph{Experimental Result} \autoref{table:model-eval} shows the translation quality of (1) MT sentences (baseline), (2) post-edited sentences with XLM-R-based translation suggestion, and (3) post-edited sentences with the proposed translation suggestion model. 
When MT sentences are post-edited based on QE prediction with the top-1 suggestion, TER and BLEU are improved over the baseline by $-$2.33 and $+$1.56, respectively. This result suggests that our QE and translation suggestion models can be used to improve MT performance without human intervention.
When the top-5 suggestions are provided, TER and BLEU are improved by $-$6.01 and $+$6.15, respectively, for the QE prediction condition and improved by $-$14.07 and $+$20.13, respectively, for the oracle QE condition.
These results imply that post-editing based on translation suggestions can significantly improve the translation quality.
Finally, the proposed model significantly outperforms XLM-R in all experimental settings, showing that fine-tuning XLM-R with the modified TLM objective is effective for the suggestion performance.

\subsection{User Study}

We conducted a user study to evaluate the effectiveness of IntelliCAT.

\paragraph{Tasks and Stimuli}
We asked participants to translate an English document to German using the given interface.
As stimuli, we prepared three English documents, each with 12 sentences and 130, 160, and 164 words.
The documents included 22, 18, and 20 styles, respectively (e.g., bold, italic, or a footnote), and participants were also asked to apply these styles in the target document.

\paragraph{Translation Interfaces}
We compared three translation interfaces: \textbf{MSWord}, \textbf{MT-Only}, and \textbf{Full}.
In \textbf{MSWord}, the participants were asked to translate documents using a popular word processor, Microsoft Word. In this baseline condition, two Microsoft Word instances were shown side-by-side: one showing an English document (source) and the other showing an empty document where one could type the translated sentences (target).
In \textbf{MT-Only}, participants started with a machine-translated document on IntelliCAT without QE, translation suggestion, and word alignment; they had to edit incorrect parts and transfer styles by themselves. In \textbf{Full}, the participants could use all the features of IntelliCAT.

\paragraph{Participants and Study Design}

We recruited nine participants (aged 23--31 years). All participants majored in German and were fluent in both English and German.
We adopted a within-subject design; each participant tested all three interfaces and three documents. 
Thus, our study consisted of nine (participants) $\times$ 3 (conditions) = 27 trials in total.
The order of interfaces and documents was counterbalanced using a $3 \times 3$ Latin square to alleviate the possible bias of learning effects or fatigue. For each trial, we measured the translation completion time.

\paragraph{Procedure}
Participants attended a training session for ten minutes, where they tried each interface with a short sample document. 
Subsequently, they performed three translation tasks with different interfaces.
We allowed them to look up words for which they did not know the translation before starting each translation task.
Upon completing the three tasks, participants responded to a system usability scale (SUS) questionnaire \citep{brooke1996sus}, and we gathered subjective feedback. The entire session took approximately 90 min per participant.

\begin{figure*}
	\centering\includegraphics[scale=0.35]{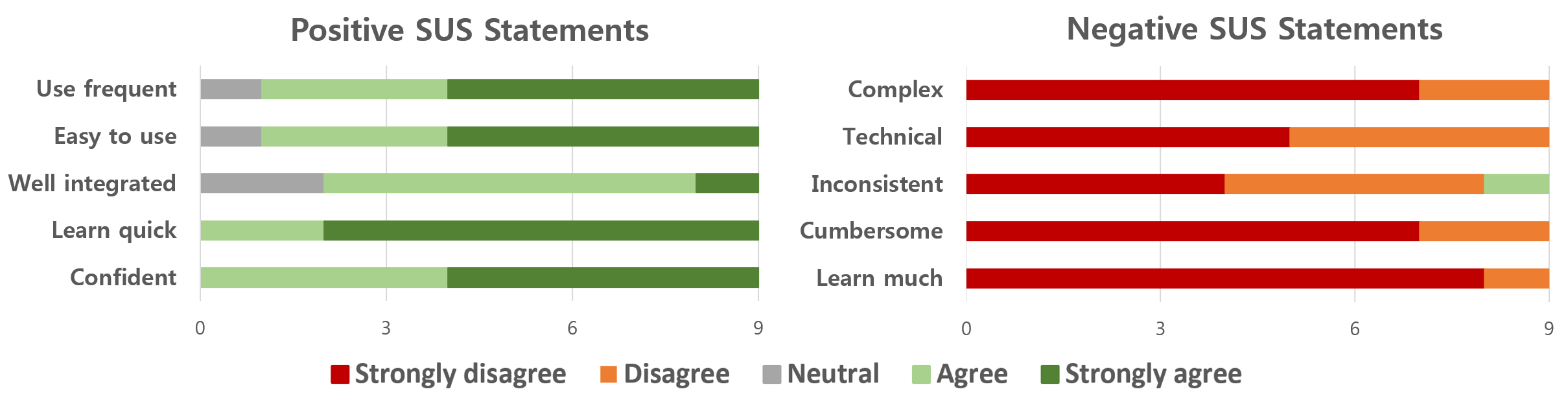}
	\caption{\label{sus-figure} SUS Feedback. The usability of IntelliCAT was evaluated as an excellent level with a score of 88.61±7.82.}
\end{figure*}

\begin{table}[h]
\centering
\begin{tabular}{lc}
\toprule
 Interface & Avg. time (s) \\
 \midrule
MSWord & 1178.78 $\pm$ 280.41 \\
MT-Only & 688.00 $\pm$ 175.02 \\
Full & \textbf{555.66 $\pm$ 200.81} \\
\bottomrule
\end{tabular}
\caption{\label{user-study-table} Translation completion time. The differences between the three interface conditions are statistically significant.}
\end{table}

\paragraph{Result and Discussion}

\autoref{user-study-table} summarizes the result of the user study. 
A repeated measures ANOVA with a Greenhouse-Geisser correction found a significant difference in completion time between the three translation interfaces ($F(1.306, 10.449) = 56.398$, $p < 0.001$).
Post hoc tests using the Bonferroni correction revealed that \textbf{All} (555.66 ± 200.81 s) was significantly faster than \textbf{MT-Only} (688.00 ± 175.02 s) ($p = 0.013$) and \textbf{MT-Only} was significantly faster than \textbf{MSWord} (1,178.78 ± 280.41 s) ($p < 0.001$).
These results suggest that our QE, translation suggestion, and word alignment features could further accelerate post-editing (a 19.2\% speedup) (\textbf{All} vs. \textbf{MT-Only}), and
our system could reduce the translation time by more than half (52.9\%) compared to translating from scratch (\textbf{All} vs. \textbf{MSWord}).

We could not find a significant difference between documents ($F(1.964, 15.712) = 0.430$, $ns$) with the same statistical procedure, which suggests that the translation difficulties of the three English documents were not statistically different.

Our interface received a mean SUS score of 88.61 ($\sigma = 7.82)$, which is slightly higher than the score for an ``Excellent'' adjective ratings (85.58, \citet{bangor2008empirical}). 
Eight out of nine participants reported that QE was useful for proofreading purposes; P2 stated, ``With QE, I could double-check the words that are possibly wrong.'' All participants evaluated the translation suggestions to be useful; P7 mentioned ``Translation suggestion was very convenient. It might significantly reduce the dependence on the dictionary.''

Overall, the user study results demonstrated the effectiveness of IntelliCAT both quantitatively and qualitatively, and we found that human translators could streamline their post-editing process with the three features provided in IntelliCAT. 

\section{Conclusion and Future Work}

In this paper, we introduce IntelliCAT, an intelligent MT post-editing interface for document translation. The interface provides three neural network-based features to assist post-editing: (1) sentence-level and word-level QEs, (2) alternative translation suggestions for words or phrases, and (3) automatic formatting of the translated document based on word alignments.
The model evaluation shows that post-editing based on the proposed QE and translation suggestion models can significantly improve the quality of translation. Moreover, the user study shows that these features significantly accelerate post-editing, achieving a 52.9\% speedup in translation time compared to translating from scratch.
Finally, the usability of IntelliCAT was evaluated as an ``excellent'' level, with a SUS score of 88.61.

In future work, we will build a pipeline that continuously improves the performance of neural models based on automatically collected triplets consisting of source, MT, and post-edited sentences. We will implement an automatic post-editing \citep{ape-20} model to continuously improve MT performance and apply online learning to QE models to continually enhance QE performance.

\bibliographystyle{acl_natbib}
\bibliography{anthology,acl2021}

\begin{thebibliography}{36}
\expandafter\ifx\csname natexlab\endcsname\relax\def\natexlab#1{#1}\fi

\bibitem[{Alabau et~al.(2014)Alabau, Buck, Carl, Casacuberta,
  Garc{\'\i}a-Mart{\'\i}nez, Germann, Gonz{\'a}lez-Rubio, Hill, Koehn, Leiva
  et~al.}]{casmacat}
Vicent Alabau, Christian Buck, Michael Carl, Francisco Casacuberta, Mercedes
  Garc{\'\i}a-Mart{\'\i}nez, Ulrich Germann, Jes{\'u}s Gonz{\'a}lez-Rubio,
  Robin Hill, Philipp Koehn, Luis~A Leiva, et~al. 2014.
\newblock Casmacat: A computer-assisted translation workbench.
\newblock In \emph{Proceedings of the Demonstrations at the 14th Conference of
  the European Chapter of the Association for Computational Linguistics}, pages
  25--28.

\bibitem[{Aranberri et~al.(2014)Aranberri, Labaka, Diaz~de Ilarraza, and
  Sarasola}]{aranberri2014comparison}
Nora Aranberri, Gorka Labaka, A~Diaz~de Ilarraza, and Kepa Sarasola. 2014.
\newblock Comparison of post-editing productivity between professional
  translators and lay users.
\newblock In \emph{Proceeding of AMTA Third Workshop on Post-editing Technology
  and Practice (WPTP-3), Vancouver, Canada}, pages 20--33.

\bibitem[{Bangor et~al.(2008)Bangor, Kortum, and Miller}]{bangor2008empirical}
Aaron Bangor, Philip~T Kortum, and James~T Miller. 2008.
\newblock An empirical evaluation of the system usability scale.
\newblock \emph{Intl. Journal of Human--Computer Interaction}, 24(6):574--594.

\bibitem[{Barrault et~al.(2020)Barrault, Biesialska, Bojar, Costa-juss{\`a},
  Federmann, Graham, Grundkiewicz, Haddow, Huck, Joanis
  et~al.}]{wmt20-findings}
Lo{\"\i}c Barrault, Magdalena Biesialska, Ond{\v{r}}ej Bojar, Marta~R
  Costa-juss{\`a}, Christian Federmann, Yvette Graham, Roman Grundkiewicz,
  Barry Haddow, Matthias Huck, Eric Joanis, et~al. 2020.
\newblock Findings of the 2020 conference on machine translation (wmt20).
\newblock In \emph{Proceedings of the Fifth Conference on Machine Translation},
  pages 1--55.

\bibitem[{Van~den Bergh et~al.(2015)Van~den Bergh, Geurts, Degraen, Haesen,
  Van~der Lek-Ciudin, Coninx et~al.}]{van2015recommendations}
Jan Van~den Bergh, Eva Geurts, Donald Degraen, Mieke Haesen, Iulianna Van~der
  Lek-Ciudin, Karin Coninx, et~al. 2015.
\newblock Recommendations for translation environments to improve
  translators’ workflows.
\newblock \emph{Translating and the Computer}, 37:106--119.

\bibitem[{Brooke(1996)}]{brooke1996sus}
John Brooke. 1996.
\newblock Sus: a “quick and dirty’usability.
\newblock \emph{Usability evaluation in industry}, 189.

\bibitem[{Chatterjee et~al.(2020)Chatterjee, Freitag, Negri, and
  Turchi}]{ape-20}
Rajen Chatterjee, Markus Freitag, Matteo Negri, and Marco Turchi. 2020.
\newblock Findings of the wmt 2020 shared task on automatic post-editing.
\newblock In \emph{Proceedings of the Fifth Conference on Machine Translation},
  pages 646--659.

\bibitem[{Conneau et~al.(2020)Conneau, Khandelwal, Goyal, Chaudhary, Wenzek,
  Guzm{\'a}n, Grave, Ott, Zettlemoyer, and Stoyanov}]{xlm-r}
Alexis Conneau, Kartikay Khandelwal, Naman Goyal, Vishrav Chaudhary, Guillaume
  Wenzek, Francisco Guzm{\'a}n, {\'E}douard Grave, Myle Ott, Luke Zettlemoyer,
  and Veselin Stoyanov. 2020.
\newblock Unsupervised cross-lingual representation learning at scale.
\newblock In \emph{Proceedings of the 58th Annual Meeting of the Association
  for Computational Linguistics}, pages 8440--8451.

\bibitem[{Coppers et~al.(2018)Coppers, Van~den Bergh, Luyten, Coninx, Van~der
  Lek-Ciudin, Vanallemeersch, and Vandeghinste}]{intellingo}
Sven Coppers, Jan Van~den Bergh, Kris Luyten, Karin Coninx, Iulianna Van~der
  Lek-Ciudin, Tom Vanallemeersch, and Vincent Vandeghinste. 2018.
\newblock Intellingo: An intelligible translation environment.
\newblock In \emph{Proceedings of the 2018 CHI Conference on Human Factors in
  Computing Systems}, pages 1--13.

\bibitem[{Do~Carmo(2020)}]{do2020comparing}
F{\'e}lix Do~Carmo. 2020.
\newblock Comparing post-editing based on four editing actions against
  translating with an auto-complete feature.
\newblock In \emph{Proceedings of the 22nd Annual Conference of the European
  Association for Machine Translation}, pages 421--430.

\bibitem[{Federico et~al.(2014)Federico, Bertoldi, Cettolo, Negri, Turchi,
  Trombetti, Cattelan, Farina, Lupinetti, Martines et~al.}]{matecat}
Marcello Federico, Nicola Bertoldi, Mauro Cettolo, Matteo Negri, Marco Turchi,
  Marco Trombetti, Alessandro Cattelan, Antonio Farina, Domenico Lupinetti,
  Andrea Martines, et~al. 2014.
\newblock The matecat tool.
\newblock In \emph{COLING (Demos)}, pages 129--132.

\bibitem[{Fonseca et~al.(2019)Fonseca, Yankovskaya, Martins, Fishel, and
  Federmann}]{qe-2019}
Erick Fonseca, Lisa Yankovskaya, Andr{\'e}~FT Martins, Mark Fishel, and
  Christian Federmann. 2019.
\newblock Findings of the wmt 2019 shared tasks on quality estimation.
\newblock In \emph{Proceedings of the Fourth Conference on Machine Translation
  (Volume 3: Shared Task Papers, Day 2)}, pages 1--10.

\bibitem[{Garg et~al.(2019)Garg, Peitz, Nallasamy, and
  Paulik}]{word-alignment-jointly-learning}
Sarthak Garg, Stephan Peitz, Udhyakumar Nallasamy, and Matthias Paulik. 2019.
\newblock Jointly learning to align and translate with transformer models.
\newblock In \emph{Proceedings of the 2019 Conference on Empirical Methods in
  Natural Language Processing and the 9th International Joint Conference on
  Natural Language Processing (EMNLP-IJCNLP)}, pages 4443--4452.

\bibitem[{Green et~al.(2014{\natexlab{a}})Green, Chuang, Heer, and
  Manning}]{lilt}
Spence Green, Jason Chuang, Jeffrey Heer, and Christopher~D Manning.
  2014{\natexlab{a}}.
\newblock Predictive translation memory: A mixed-initiative system for human
  language translation.
\newblock In \emph{Proceedings of the 27th annual ACM symposium on User
  interface software and technology}, pages 177--187.

\bibitem[{Green et~al.(2013)Green, Heer, and Manning}]{green2013efficacy}
Spence Green, Jeffrey Heer, and Christopher~D Manning. 2013.
\newblock The efficacy of human post-editing for language translation.
\newblock In \emph{Proceedings of the SIGCHI conference on human factors in
  computing systems}, pages 439--448.

\bibitem[{Green et~al.(2014{\natexlab{b}})Green, Wang, Chuang, Heer, Schuster,
  and Manning}]{green2014human}
Spence Green, Sida~I Wang, Jason Chuang, Jeffrey Heer, Sebastian Schuster, and
  Christopher~D Manning. 2014{\natexlab{b}}.
\newblock Human effort and machine learnability in computer aided translation.
\newblock In \emph{Proceedings of the 2014 Conference on Empirical Methods in
  Natural Language Processing (EMNLP)}, pages 1225--1236.

\bibitem[{Herbig et~al.(2020)Herbig, D{\"u}wel, Pal, Meladaki, Monshizadeh,
  Kr{\"u}ger, and van Genabith}]{mmpe}
Nico Herbig, Tim D{\"u}wel, Santanu Pal, Kalliopi Meladaki, Mahsa Monshizadeh,
  Antonio Kr{\"u}ger, and Josef van Genabith. 2020.
\newblock Mmpe: A multi-modal interface for post-editing machine translation.
\newblock In \emph{Proceedings of the 58th Annual Meeting of the Association
  for Computational Linguistics}, pages 1691--1702.

\bibitem[{Joshi et~al.(2020)Joshi, Chen, Liu, Weld, Zettlemoyer, and
  Levy}]{spanbert}
Mandar Joshi, Danqi Chen, Yinhan Liu, Daniel~S Weld, Luke Zettlemoyer, and Omer
  Levy. 2020.
\newblock Spanbert: Improving pre-training by representing and predicting
  spans.
\newblock \emph{Transactions of the Association for Computational Linguistics},
  8:64--77.

\bibitem[{Klein et~al.(2017)Klein, Kim, Deng, Senellart, and Rush}]{opennmt}
Guillaume Klein, Yoon Kim, Yuntian Deng, Jean Senellart, and Alexander~M. Rush.
  2017.
\newblock \href {https://doi.org/10.18653/v1/P17-4012} {Open{NMT}: Open-source
  toolkit for neural machine translation}.
\newblock In \emph{Proc. ACL}.

\bibitem[{Koehn(2009)}]{koehn2009process}
Philipp Koehn. 2009.
\newblock A process study of computer-aided translation.
\newblock \emph{Machine Translation}, 23(4):241--263.

\bibitem[{Kudo(2018)}]{subword-regularization}
Taku Kudo. 2018.
\newblock Subword regularization: Improving neural network translation models
  with multiple subword candidates.
\newblock In \emph{Proceedings of the 56th Annual Meeting of the Association
  for Computational Linguistics (Volume 1: Long Papers)}, pages 66--75.

\bibitem[{Lample and Conneau(2019)}]{xlm}
Guillaume Lample and Alexis Conneau. 2019.
\newblock Cross-lingual language model pretraining.
\newblock \emph{arXiv preprint arXiv:1901.07291}.

\bibitem[{Lee(2020)}]{my-qe}
Dongjun Lee. 2020.
\newblock \href {https://www.aclweb.org/anthology/2020.wmt-1.118} {Two-phase
  cross-lingual language model fine-tuning for machine translation quality
  estimation}.
\newblock In \emph{Proceedings of the Fifth Conference on Machine Translation},
  pages 1024--1028, Online. Association for Computational Linguistics.

\bibitem[{Lee et~al.(2020)Lee, Lee, Shin, Jung, Kim, and Lee}]{ape-2}
Jihyung Lee, WonKee Lee, Jaehun Shin, Baikjin Jung, Young-Gil Kim, and
  Jong-Hyeok Lee. 2020.
\newblock Postech-etri’s submission to the wmt2020 ape shared task: Automatic
  post-editing with cross-lingual language model.
\newblock In \emph{Proceedings of the Fifth Conference on Machine Translation},
  pages 777--782.

\bibitem[{Lo and Joanis(2020)}]{lo2020improving}
Chi-kiu Lo and Eric Joanis. 2020.
\newblock Improving parallel data identification using iteratively refined
  sentence alignments and bilingual mappings of pre-trained language models.
\newblock In \emph{Proceedings of the Fifth Conference on Machine Translation},
  pages 972--978.

\bibitem[{Moorkens and O’Brien(2017)}]{moorkens2017assessing}
Joss Moorkens and Sharon O’Brien. 2017.
\newblock Assessing user interface needs of post-editors of machine
  translation.
\newblock \emph{Human issues in translation technology}, pages 109--130.

\bibitem[{Och and Ney(2003)}]{giza}
Franz~Josef Och and Hermann Ney. 2003.
\newblock A systematic comparison of various statistical alignment models.
\newblock \emph{Computational Linguistics}, 29(1):19--51.

\bibitem[{Pal et~al.(2016)Pal, Zampieri, Naskar, Nayak, Vela, and van
  Genabith}]{catalog_online}
Santanu Pal, Marcos Zampieri, Sudip~Kumar Naskar, Tapas Nayak, Mihaela Vela,
  and Josef van Genabith. 2016.
\newblock Catalog online: Porting a post-editing tool to the web.
\newblock In \emph{Proceedings of the Tenth International Conference on
  Language Resources and Evaluation (LREC'16)}, pages 599--604.

\bibitem[{Papineni et~al.(2002)Papineni, Roukos, Ward, and Zhu}]{bleu}
Kishore Papineni, Salim Roukos, Todd Ward, and Wei-Jing Zhu. 2002.
\newblock \href {https://doi.org/10.3115/1073083.1073135} {{B}leu: a method for
  automatic evaluation of machine translation}.
\newblock In \emph{Proceedings of the 40th Annual Meeting of the Association
  for Computational Linguistics}, pages 311--318, Philadelphia, Pennsylvania,
  USA. Association for Computational Linguistics.

\bibitem[{Santy et~al.(2019)Santy, Dandapat, Choudhury, and Bali}]{inmt}
Sebastin Santy, Sandipan Dandapat, Monojit Choudhury, and Kalika Bali. 2019.
\newblock Inmt: Interactive neural machine translation prediction.
\newblock In \emph{Proceedings of the 2019 Conference on Empirical Methods in
  Natural Language Processing and the 9th International Joint Conference on
  Natural Language Processing (EMNLP-IJCNLP): System Demonstrations}, pages
  103--108.

\bibitem[{Sennrich et~al.(2016)Sennrich, Haddow, and Birch}]{bpe}
Rico Sennrich, Barry Haddow, and Alexandra Birch. 2016.
\newblock Neural machine translation of rare words with subword units.
\newblock In \emph{Proceedings of the 54th Annual Meeting of the Association
  for Computational Linguistics (Volume 1: Long Papers)}, pages 1715--1725.

\bibitem[{Snover et~al.(2006)Snover, Dorr, Schwartz, Micciulla, and
  Makhoul}]{ter-tool}
Matthew Snover, Bonnie Dorr, Richard Schwartz, Linnea Micciulla, and John
  Makhoul. 2006.
\newblock A study of translation edit rate with targeted human annotation.
\newblock In \emph{Proceedings of association for machine translation in the
  Americas}, volume 200.

\bibitem[{Specia et~al.(2020)Specia, Blain, Fomicheva, Fonseca, Chaudhary,
  Guzm{\'e}n, and Martins}]{qe-20}
Lucia Specia, Fr{\'e}d{\'e}ric Blain, Marina Fomicheva, Erick Fonseca, Vishrav
  Chaudhary, Francisco Guzm{\'e}n, and Andr{\'e} F.~T. Martins. 2020.
\newblock \href {https://www.aclweb.org/anthology/2020.wmt-1.79} {Findings of
  the wmt 2020 shared task on quality estimation}.
\newblock In \emph{Proceedings of the Fifth Conference on Machine Translation},
  pages 743--764, Online. Association for Computational Linguistics.

\bibitem[{Toral et~al.(2018)Toral, Wieling, and Way}]{toral2018post}
Antonio Toral, Martijn Wieling, and Andy Way. 2018.
\newblock Post-editing effort of a novel with statistical and neural machine
  translation.
\newblock \emph{Frontiers in Digital Humanities}, 5:9.

\bibitem[{Vaswani et~al.(2017)Vaswani, Shazeer, Parmar, Uszkoreit, Jones,
  Gomez, Kaiser, and Polosukhin}]{transformer}
Ashish Vaswani, Noam Shazeer, Niki Parmar, Jakob Uszkoreit, Llion Jones,
  Aidan~N Gomez, {\L}ukasz Kaiser, and Illia Polosukhin. 2017.
\newblock Attention is all you need.
\newblock In \emph{Advances in neural information processing systems}, pages
  5998--6008.

\bibitem[{Wang et~al.(2020)Wang, Wang, Fan, Zhang, Lu, Ge, Shi, and
  Zhao}]{ape-1}
Jiayi Wang, Ke~Wang, Kai Fan, Yuqi Zhang, Jun Lu, Xin Ge, Yangbin Shi, and
  Yu~Zhao. 2020.
\newblock Alibaba’s submission for the wmt 2020 ape shared task: Improving
  automatic post-editing with pre-trained conditional cross-lingual bert.
\newblock In \emph{Proceedings of the Fifth Conference on Machine Translation},
  pages 789--796.

\end{thebibliography}

\appendix
\section{Sample Document Translation}
\label{sec:appendix-a}
\begin{figure*}[h]
	\centering\includegraphics[scale=0.26]{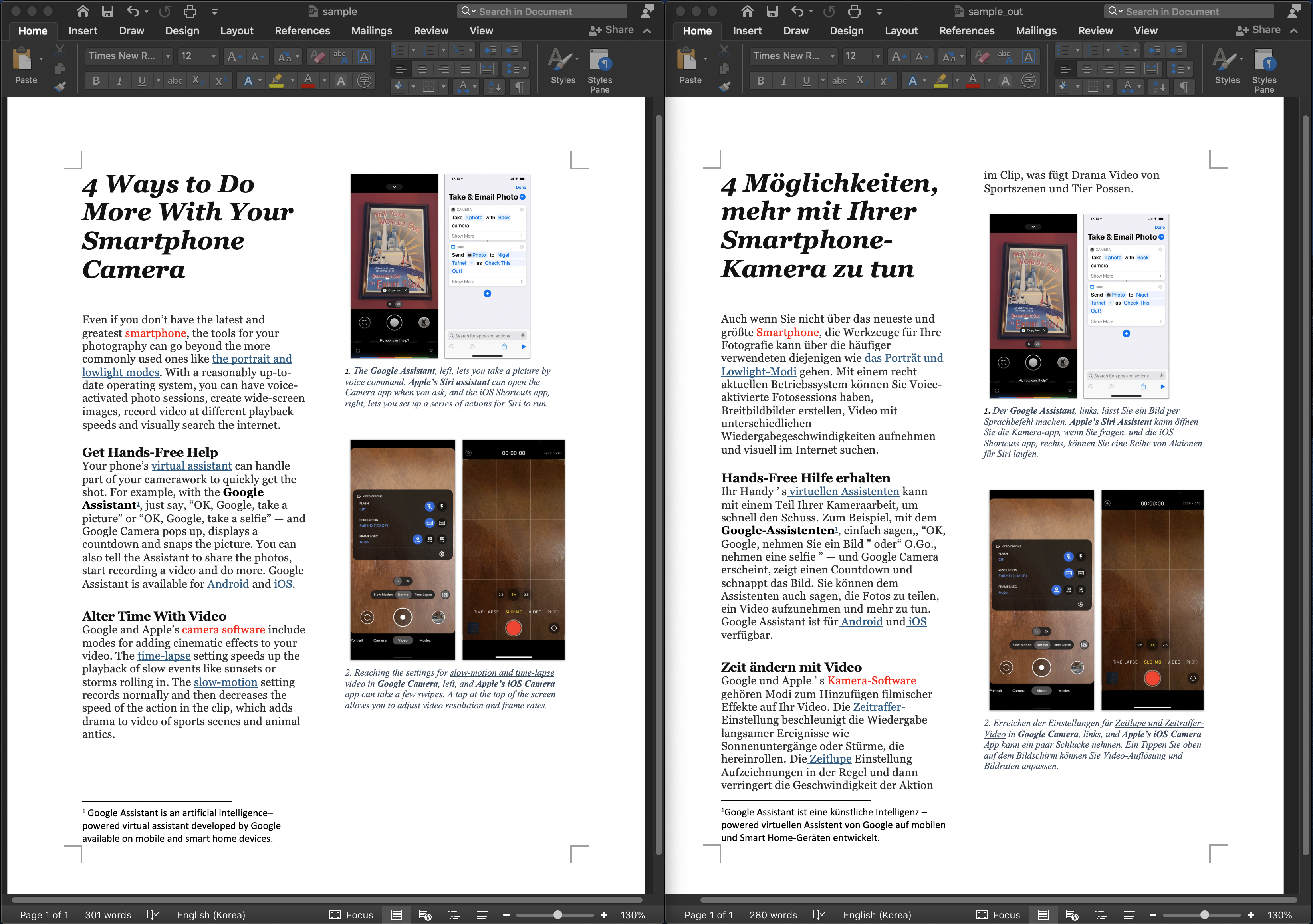}
	\caption{\label{screenshot-sample-doc} A sample document (left) and the translated document (right) without human intervention.}
\end{figure*}

\autoref{screenshot-sample-doc} shows a sample document and the translated document using IntelliCAT without human intervention.

\end{document}